\documentclass{article}
\usepackage{iclr2027_conference,times}

\usepackage{amsmath,amsfonts,bm}

\def\eqref#1{equation~\ref{#1}}

\def\1{\bm{1}}

\DeclareMathAlphabet{\mathsfit}{\encodingdefault}{\sfdefault}{m}{sl}
\SetMathAlphabet{\mathsfit}{bold}{\encodingdefault}{\sfdefault}{bx}{n}

\usepackage{hyperref}
\usepackage{url}
\usepackage{booktabs}
\usepackage{array}
\usepackage{colortbl}
\usepackage{multirow}
\usepackage{graphicx}
\usepackage{enumitem}
\usepackage{float}
\usepackage{caption}

\floatstyle{ruled}
\newfloat{algorithm}{tbp}{loa}
\floatname{algorithm}{Algorithm}
\newcommand{\score}[2]{\makebox[3.0em][r]{#1}/\makebox[3.0em][l]{#2}}

\title{Exploration via Activation-Space Steering for Reinforcement Learning with Verifiable Rewards}

\author{Yunan Zeng\textsuperscript{1} \quad
Jinhao Jing\textsuperscript{2} \quad
Xi Cheng\textsuperscript{3} \quad
Yuwang Wang\textsuperscript{4}\thanks{Corresponding author: \texttt{wang-yuwang@tsinghua.edu.cn}.} \\
\textnormal{\textsuperscript{1}Fudan University} \\
\textnormal{\textsuperscript{2}The Chinese University of Hong Kong, Shenzhen} \\
\textnormal{\textsuperscript{3}Kunlun Inc. \quad \textsuperscript{4}Tsinghua University} \\
{\small\normalfont\texttt{ynzeng24@m.fudan.edu.cn}}}

\iclrfinalcopy
\begin{document}
\maketitle
\fancyhead[L]{}

\begin{abstract}
Reinforcement learning with verifiable rewards (RLVR) trains a language model by sampling responses, scoring them with an automatic verifier, and updating the policy from the resulting rewards. The policy therefore receives direct learning signals only from trajectories that appear in its rollouts. Yet repeated sampling from the same policy can concentrate on similar reasoning paths, leaving useful alternatives unexplored under a fixed rollout budget. We introduce Activation-Space Exploration (ASE), an online method for diversifying training rollouts. By clustering activation differences between correct and incorrect responses to the same problem, ASE builds a structured bank of latent exploration vectors. Different vectors are then applied to subsequent rollouts to explore alternative reasoning paths. On Qwen3-1.7B-Base and Qwen3-8B-Base trained on DAPO-Math-17k, ASE improves average Mean@8 across six mathematical reasoning benchmarks over GRPO by 4.87 and 1.87 percentage points, respectively, under the same rollout budget.
\end{abstract}

\section{Introduction}
Reinforcement learning improves a model by repeatedly sampling behaviors, evaluating their outcomes with rewards, and increasing the probability of behaviors that receive higher rewards~\citep{sutton2018reinforcement}. Reinforcement learning with verifiable rewards (RLVR) makes this process scalable by using automated verifiers to assign outcome-based rewards to generated responses, such as mathematical solutions and programs~\citep{le2022coderl,shao2024deepseekmath,yu2025dapo}. A fundamental constraint, however, is that the policy can only learn from behaviors that appear in its sampled rollouts~\citep{shao2024deepseekmath,yue2025rlvr}. Consequently, policy optimization is shaped by \textbf{which trajectories the rollout procedure makes available for learning}, a dependence that creates an exploration bottleneck under a finite rollout budget. Standard GRPO independently samples multiple trajectories from the same current policy~\citep{shao2024deepseekmath}. Increasing the number of rollouts provides more Monte Carlo samples, but leaves the underlying proposal distribution unchanged. As the policy concentrates probability mass on a limited set of reasoning patterns, multiple samples may repeatedly visit similar solution modes, reproduce the same intermediate steps, or make correlated errors. Useful alternatives that occupy low-probability regions of the policy can therefore remain rarely observed and receive little direct reward feedback. The bottleneck is thus not merely how many trajectories are sampled, but \textbf{how a fixed rollout budget is distributed across the policy's reasoning space}.

Several approaches seek to use a finite RLVR rollout budget more effectively. Adaptive sampling allocates more rollouts to difficult problems~\citep{liao2025enhancing}, while data filtering and curricula prioritize more informative training problems~\citep{yu2025dapo,chen2025selfevolving}. These methods change where sampling effort is spent, rather than how trajectories for the same problem are generated. Adjusting decoding temperature increases token-level variation~\citep{dang2026temperature}, but does not specify which alternative reasoning patterns to explore. Replay reuses previously observed trajectories~\citep{li2025repo}, while novelty rewards and high-entropy prefix regeneration diversify generation around observed paths~\citep{hu2026extra}. Together, these methods improve how rollouts are allocated, change how randomly tokens are drawn, and reuse previously generated trajectories. They leave the model's internal representations largely unexplored as a way to construct different rollout proposals for the same prompt.

We investigate activation-space exploration, where temporary interventions on hidden representations are used to construct multiple rollout proposal distributions while leaving both the problem context and model parameters unchanged. Activation steering has shown that directional interventions on Transformer representations can systematically alter model behavior and generation trajectories~\citep{beaglehole2026universal,chen2025persona}. Rather than using such interventions as inference-time behavior controllers, we ask whether they can serve as a training-time exploration mechanism: different latent directions can perturb generation toward different regions of the policy's reasoning space, exposing the optimizer to useful trajectories that ordinary sampling may rarely encounter.

\begin{figure*}[t]
\centering
\includegraphics[width=0.6\linewidth]{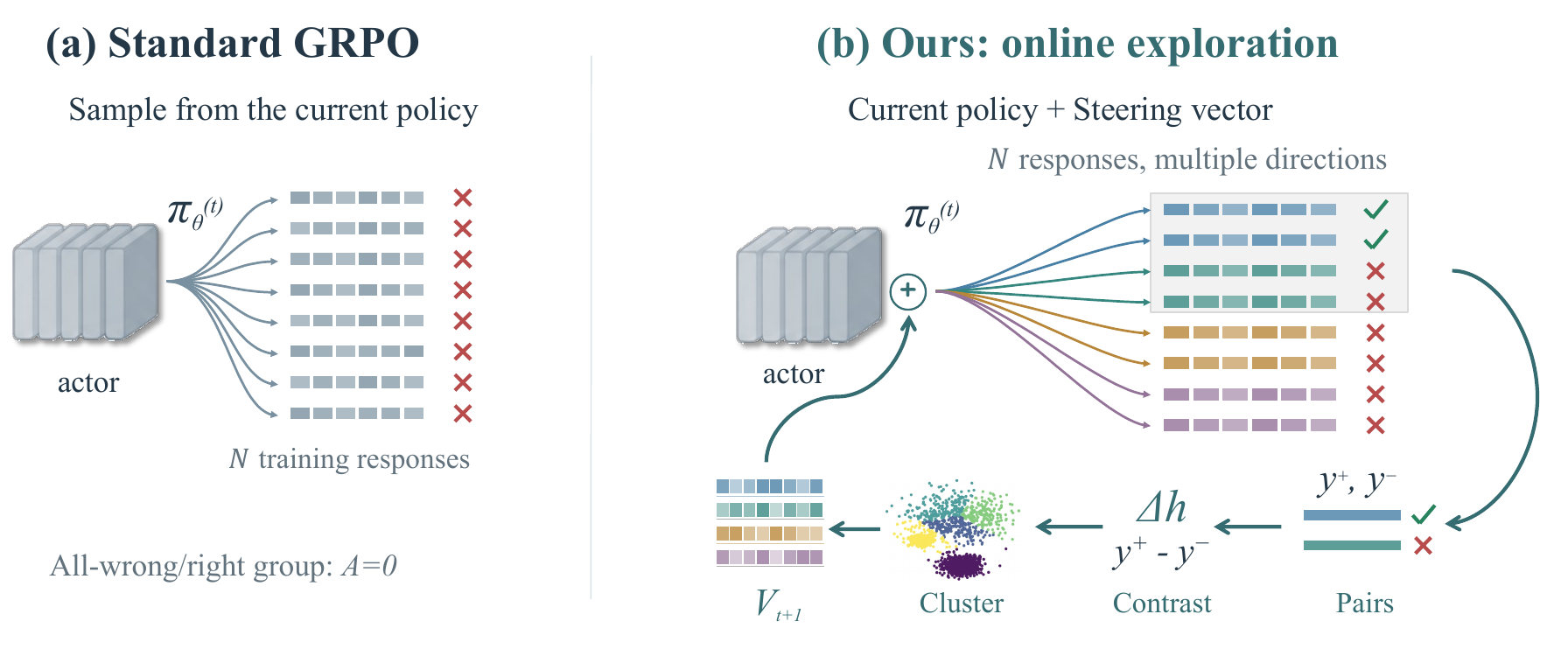}
\caption{Motivation and high-level overview of activation-space exploration. Standard GRPO samples all responses from one proposal distribution, whereas ASE uses outcome feedback to maintain multiple activation-space directions, routes responses through different proposals, and updates the actor and vector bank online.}
\label{fig:overview}
\end{figure*}

Based on this idea, we propose \textbf{Activation-Space Exploration (ASE)}, an online exploration mechanism for RLVR that learns where to explore from the policy's own verifier results. For each problem, successful and unsuccessful trajectories generated during training provide within-problem contrasts that isolate activation differences associated with task outcome while controlling for problem content. We find that the activation contrasts exhibit heterogeneous structure; we therefore cluster and aggregate them into multiple exploration directions maintained in an online vector bank. During subsequent rollout generation, different trajectories for the same problem are routed through different latent directions, producing a collection of steering-conditioned proposal distributions rather than repeatedly sampling from a single one (Figure~\ref{fig:overview}). Stochastic decoding is retained within every route, so ASE complements rather than replaces ordinary token sampling. Verifier rewards from the steered rollouts drive both subsequent exploration updates and optimization of the base policy, leaving evaluation and deployment intervention-free.

We evaluate ASE on Qwen3-1.7B-Base~\citep{qwen3technicalreport} and Qwen3-8B-Base~\citep{qwen3technicalreport} using DAPO-Math-17k~\citep{yu2025dapo} under the same rollout budget as standard GRPO. All activation interventions are disabled during evaluation. Across six mathematical reasoning benchmarks, ASE improves average Mean@8 over GRPO by \textbf{4.87 percentage points on Qwen3-1.7B-Base and 1.87 points on Qwen3-8B-Base}, while also transferring to out-of-distribution reasoning benchmarks. These results show that activation-space exploration during training can improve the learned policy across model scales and benchmark distributions under a fixed rollout budget.


\section{Related Work}
\paragraph{Reinforcement learning with verifiable rewards.}
RLVR is now a major paradigm for improving mathematical and code reasoning in language models. \citet{shao2024deepseekmath} introduced GRPO, which constructs policy-gradient signals from relative rewards among multiple trajectories for the same problem; subsequent work studied the stability and efficiency of scaling this formulation~\citep{yu2025dapo}. RLVR gains depend strongly on trajectories collected by the current policy. Policy optimization may mostly redistribute probability mass toward high-reward paths already available to the base model and can reduce solution-space coverage under large-scale sampling~\citep{yue2025rlvr}; policy entropy can also decrease continuously and eventually limit further exploration~\citep{cui2025entropy}. At the token level, a small number of high-entropy branching positions have a disproportionate effect on reasoning paths~\citep{wang2025highentropy}. Recent methods encourage flatter distributions and answer diversity in the objective~\citep{li2026edo}, or combine trajectory novelty rewards with high-entropy prefix regeneration to discover different correct solutions~\citep{hu2026extra}. RePO retrieves historical off-policy trajectories from a replay buffer and estimates their advantages separately from current on-policy rollouts~\citep{li2025repo}; GRESO uses reward dynamics across epochs to probabilistically skip prompts expected to remain zero-variance~\citep{zheng2025greso}. These methods improve exploration through objectives, reward shaping, explicit trajectory operations, or data scheduling. We instead use historical rollout outcomes to construct multiple directional proposals in activation space, directly changing the sampling distribution of later training trajectories without an additional diversity reward or a modified GRPO objective.

\paragraph{Activation steering.}
Activation steering builds on the observation that high-level behaviors have controllable directions in language-model representations. Representation engineering and inference-time intervention show that identifying and modifying intermediate representations can change truthfulness, safety, and other behaviors~\citep{zou2023repe,li2023iti}. More systematic studies have verified transferable and compositional linear concept representations across hundreds of concepts in language, vision-language, and reasoning models, and demonstrated their use for steering and monitoring~\citep{beaglehole2026universal}. Activation Addition (ActAdd) constructs a direction from activation differences of contrastive prompts~\citep{turner2023actadd}; Contrastive Activation Addition (CAA) pairs examples with opposite behavior labels, averages residual-stream differences, and adds the result during generation~\citep{rimsky2024steering}. Later work extends this idea to instruction following~\citep{stolfo2025instruction} and reasoning behaviors such as uncertainty, backtracking, and hypothesis testing~\citep{venhoff2025reasoning}. Other methods automatically construct steering vectors from labeled data~\citep{cui2025painless}, or optimize a small number of trainable vectors with RL so that they carry reasoning gains~\citep{sinii2025smallvectors}. These methods generally treat a steering vector as a behavior controller applied at inference time and rely on predefined attributes, externally constructed contrast sets, or optimization of the vector itself. We reposition activation steering as a temporary training-time exploration mechanism: directions are estimated online from success and failure trajectories of the current policy, several directions form heterogeneous rollout proposals, and GRPO absorbs useful behaviors into the base policy parameters. The vector bank is discarded after training.

\section{Method}
\subsection{Overview}
Let $x\sim\mathcal D$ be a problem and let $\pi_\theta$ be a language-model policy. For a fixed $x$, standard GRPO independently samples $N$ trajectories:
\begin{equation}
y_i^{(t)}\sim\pi_{\theta^{(t)}}(\cdot\mid x),\qquad i=1,\ldots,N.
\end{equation}
With a fixed policy and sampling configuration, all trajectories are generated under the same proposal distribution. Increasing $N$ gives more Monte Carlo samples while leaving the proposal distribution unchanged. We use historical rollout trajectories to construct structured activation-space exploration directions.

These exploration directions form a steering vector bank that is updated across iterations. At iteration $t$,
\begin{equation}
\mathcal V^{(t)}=\{\mathbf v_k^{(t)}\in\mathbb R^d\}_{k=1}^{K},\qquad \|\mathbf v_k^{(t)}\|_2=1.
\end{equation}
Here $d$ is the hidden dimension of the target Transformer block and $K$ is the number of exploration routes. Each vector is a smoothed direction obtained from mean activation differences between correct and incorrect trajectories in a cluster. The vector bank is a non-parametric training state independent of actor parameters and is estimated directly from historical rollouts.

During rollout, trajectories for the same problem are assigned different routes. For route $s_i\in\{1,\ldots,K\}$, we add $\alpha\mathbf v_{s_i}^{(t)}$ to the output of the target Transformer block during generation. The scalar $\alpha$ controls the intervention strength. We also include a clean route with no activation intervention, represented by $\mathbf v_0^{(t)}=\mathbf 0$. This changes hidden states and subsequent token distributions, yielding a route-specific proposal distribution:
\begin{equation}
y_i^{(t)}\sim
\begin{cases}
\pi_{\theta^{(t)}}(\cdot\mid x), & s_i=0,\\
\widetilde\pi_{\theta^{(t)},\mathbf v_{s_i}^{(t)}}(\cdot\mid x), & s_i\in\{1,\ldots,K\}.
\end{cases}
\end{equation}
Token sampling remains stochastic within each route, and the main configuration allocates the group budget uniformly across the clean route and the $K$ steering-guided proposals. If $\mathcal V^{(t)}=\emptyset$, the rollout distribution reduces to the unsteered policy $\pi_{\theta^{(t)}}(\cdot\mid x)$.

The verifier supplies a reward $r_i^{(t)}$ after generation. GRPO converts group rewards into relative advantages and updates the base policy with these routed trajectories. Steering changes only the proposal distribution during rollout, allowing training to discover candidates rarely covered by ordinary sampling. GRPO then increases the probability of useful behaviors under the base policy, making the exploratory experience persistent. Validation and deployment disable steering, so the resulting policy can reason independently.

\begin{figure*}[t]
\centering
\includegraphics[width=0.9\linewidth]{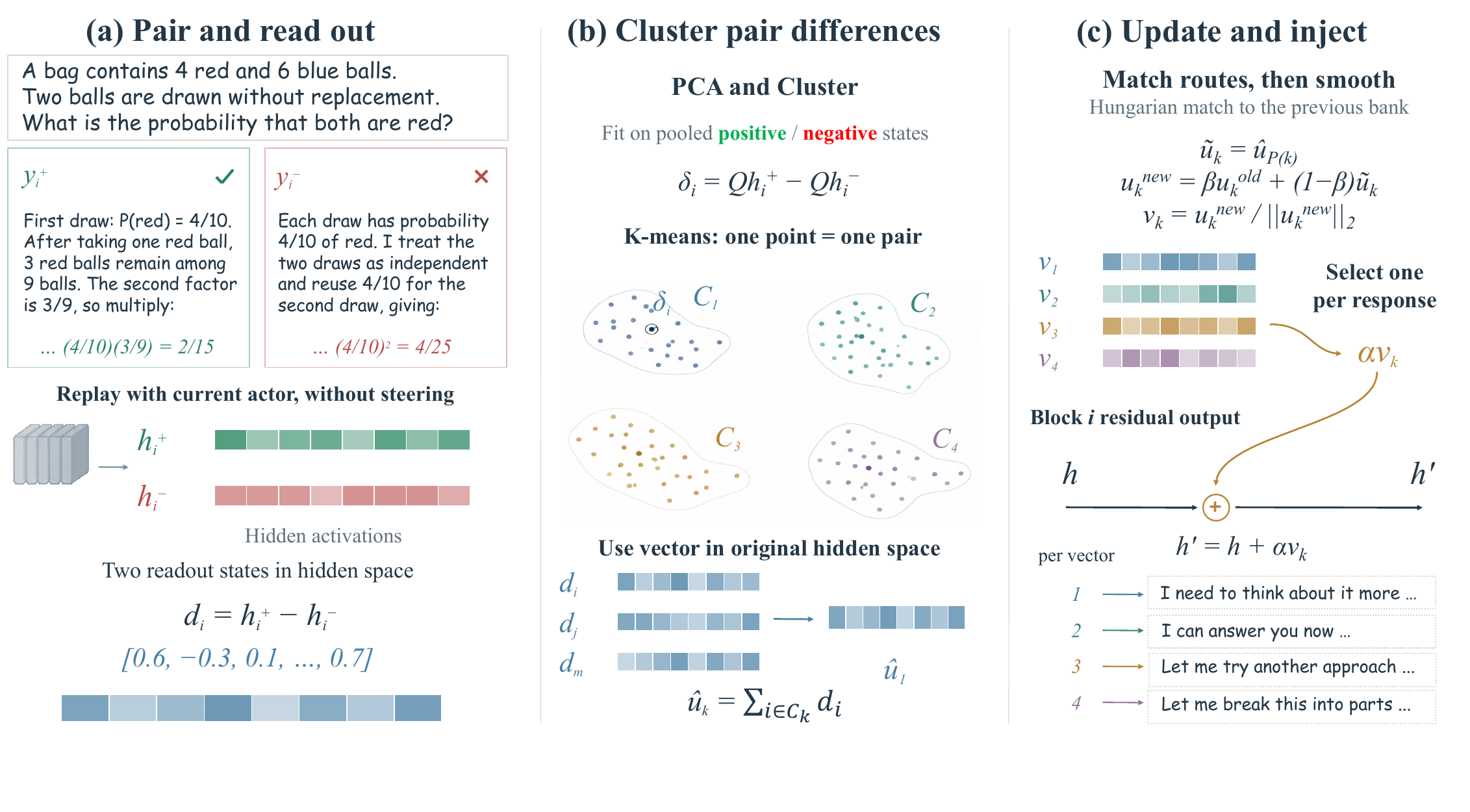}
\caption{ASE pipeline. We replay successful and unsuccessful responses to read out hidden activations, cluster their pairwise differences to form exploration directions, match and smooth the directions across updates, and inject one selected direction into the residual stream for each routed response.}
\label{fig:pipeline}
\end{figure*}

The same verifier-scored trajectories update future exploration. We construct positive--negative pairs to estimate $\mathcal V^{(t+1)}$; a newly estimated bank becomes active only after actor optimization and weight synchronization. The resulting loop is structured exploration, outcome evaluation, policy learning, and exploration update (Figure~\ref{fig:pipeline}). Section~\ref{sec:vector-bank} describes vector-bank estimation and Section~\ref{sec:rollout-optimization} describes steering-guided rollout optimization.

\subsection{Online Exploration Vector Bank}
\label{sec:vector-bank}
\paragraph{Within-problem positive--negative pairing.}
For each problem, we compare trajectories with positive and negative rewards. Positive trajectories are successful samples from the current policy, while negative trajectories provide unsuccessful references. Let $\mathcal P=\{i:r_i=1\}$ and $\mathcal N=\{i:r_i=0\}$. Only groups containing both reward types provide a within-problem contrast. For each such group, we randomly select and pair $n_{\mathrm{pair}}=\min(|\mathcal P|,|\mathcal N|)$ trajectories from the two sides. Groups whose rewards are all one or all zero do not contribute to the current steering state.

\paragraph{Activation readout.}
We replay each selected full prompt--response sequence with the current actor without steering. Let $\mathbf h_i=\phi_{\ell,p}(x,y_i)\in\mathbb R^d$ be the activation selected by position function $p$ from the output of Transformer block $\ell$. Each within-problem pair gives a raw activation difference $\mathbf d_j=\mathbf h^+_j-\mathbf h^-_j$ for $j=1,\ldots,n_{\mathrm{pair}}$.
Because the paired trajectories share the same prompt and use the same readout position rule $p$, this difference removes much of the variation caused by problem content and difficulty and provides a basis for structured exploration directions.

Paired activations accumulate across iterations in a buffer $\mathcal B$. Let $B_{\min}$ be the minimum number of pairs required for an update, $\Delta$ the minimum interval between updates, and $t_{\mathrm{last}}$ the iteration index of the most recent vector-bank update. When $|\mathcal B|\ge B_{\min}$ and $t-t_{\mathrm{last}}\ge\Delta$, we re-estimate the vector bank from the full buffer and then clear it. The readout layer, position function, $B_{\min}$, and $\Delta$ are implementation hyperparameters.

\paragraph{Difference clustering.}
A single global mean difference would compress diverse exploration patterns into one direction. These patterns may reflect variation in early strategy choice, the sequencing of intermediate steps, or later checking and revision. We therefore cluster in a dimension-reduced pair-difference space.

Suppose the buffer contains $M$ pairs $\{(\mathbf h_b^+,\mathbf h_b^-)\}_{b=1}^M$ collected across problems. To stabilize clustering when the number of pairs is limited relative to the hidden dimension, we fit a PCA projection $\mathbf Q\in\mathbb R^{d_{\mathrm{PCA}}\times d}$ on the pooled positive and negative activations and compute each pair's difference in this shared subspace as $\boldsymbol\delta_b=\mathbf Q\mathbf h_b^+-\mathbf Q\mathbf h_b^-$. Here $d_{\mathrm{PCA}}$ denotes the number of retained principal components. K-means partitions $\{\boldsymbol\delta_b\}_{b=1}^M$ into $K$ clusters $\{\mathcal C_k^{(t+1)}\}_{k=1}^K$. The set $\mathcal C_k^{(t+1)}$ contains the indices of all pairs assigned to cluster $k$.
PCA only defines the clustering space. The final direction is the within-cluster mean difference in the original hidden space:
\begin{equation}
\widehat{\mathbf u}_k^{(t+1)}=\frac{1}{|\mathcal C_k^{(t+1)}|}\sum_{b\in\mathcal C_k^{(t+1)}}(\mathbf h_b^+-\mathbf h_b^-).
\end{equation}
The clusters construct multiple local rollout proposals; we do not assume that they correspond to discrete, stable, or human-namable semantics.

\paragraph{Cross-iteration stabilization.}
Let $\mathbf u_k^{(t)}$ denote the unnormalized, temporally smoothed direction, with $\mathbf v_k^{(t)}=\mathbf u_k^{(t)}/\|\mathbf u_k^{(t)}\|_2$.
A finite buffer is affected by rollout noise, so we use an EMA with coefficient $\beta\in[0,1)$ to smooth updates. Since K-means cluster indices have no fixed meaning across updates, we first align the new candidate directions with the existing bank by solving
\begin{equation}
P^{(t+1)}=\arg\max_{P\in\mathfrak S_K}\sum_{k=1}^K\cos(\mathbf u_k^{(t)},\widehat{\mathbf u}_{P(k)}^{(t+1)}),
\end{equation}
where $\mathfrak S_K$ is the set of permutations. We use the Hungarian algorithm, then update
\begin{equation}
\mathbf u_k^{(t+1)}=\beta\mathbf u_k^{(t)}+(1-\beta)\widehat{\mathbf u}_{P^{(t+1)}(k)}^{(t+1)}.
\end{equation}
For the first update, we set $\mathbf u_k^{(t+1)}=\widehat{\mathbf u}_k^{(t+1)}$. Finally, we normalize $\mathbf u_k^{(t+1)}$ to unit $L_2$ norm to obtain $\mathbf v_k^{(t+1)}$.
The new bank is staged until actor optimization and rollout weight synchronization finish.

\subsection{Steering-Guided Rollout Optimization}
\label{sec:rollout-optimization}
\paragraph{Routed activation steering.}
When $\mathcal V^{(t)}\ne\emptyset$, we assign routes $s_i\in\{0,\ldots,K\}$ to trajectories for each problem, where $s_i=0$ is the clean route. Our main method uses all $K+1$ routes uniformly within each group. For any target-layer output $\mathbf h^{(\ell)}$ on a trajectory assigned to a steering route $s_i\in\{1,\ldots,K\}$, the intervention is
\begin{equation}
\widetilde{\mathbf h}^{(\ell)}=\mathbf h^{(\ell)}+\alpha\mathbf v_{s_i}^{(t)}.
\end{equation}
For the clean route, the output is unchanged. The intervention is applied to every forward pass of the target layer, including prompt prefill and autoregressive decoding. Since each vector has unit $L_2$ norm, $\alpha$ directly controls the injected residual. Steering is executed by the rollout engine with fixed actor parameters and is outside backpropagation.

\paragraph{GRPO update.}
For the $N$ rollouts of a fixed problem $x$, let $\mu$ and $\sigma$ be the mean and sample standard deviation of their verifier rewards. The group-relative advantage is $A_i=(r_i-\mu)/(\sigma+\epsilon)$, where $\epsilon>0$ stabilizes normalization.
We update the base policy using trajectories from both clean and steered rollout routes, and apply a bounded token-level importance weight in the clipped GRPO objective. For each response token,
\begin{equation}
\rho_{i,\tau}(\theta)=\frac{\pi_\theta(y_{i,\tau}\mid x,y_{i,<\tau})}{\pi_{\theta^{(t)}}(y_{i,\tau}\mid x,y_{i,<\tau})}.
\end{equation}
For $s_i\in\{0,\ldots,K\}$, the truncated importance ratio is
\begin{equation}
w_{i,\tau}=\min\!\left(c_{\mathrm{IS}},\frac{\pi_{\theta^{(t)}}(y_{i,\tau}\mid x,y_{i,<\tau})}{\widetilde\pi_{\theta^{(t)},\mathbf v_{s_i}^{(t)}}(y_{i,\tau}\mid x,y_{i,<\tau})}\right).
\end{equation}
Here $c_{\mathrm{IS}}>0$ caps the importance weight. This weight is computed separately at each response position. The token-level objective is
\begin{equation}
\mathcal L(\theta)=-\mathbb E_{x,i,\tau}\!\left[w_{i,\tau}\min\left(\rho_{i,\tau}(\theta)A_i,\operatorname{clip}(\rho_{i,\tau}(\theta),1-\varepsilon,1+\varepsilon)A_i\right)\right].
\end{equation}
The expectation averages over generated response tokens, and $\varepsilon>0$ sets the PPO clipping range.
Minimizing this objective updates the base policy from $\theta^{(t)}$ to $\theta^{(t+1)}$. The verifier and advantage estimator follow GRPO. The bounded importance weight adjusts the contribution of each sampled token according to its likelihood under the base policy and its rollout route. The complete training procedure is given in Algorithm~\ref{alg:ase}.

\section{Experiments}
\subsection{Experimental Setup}
\paragraph{Models and datasets.}
We train Qwen3-1.7B-Base and Qwen3-8B-Base~\citep{qwen3technicalreport} with DAPO-Math-17k~\citep{yu2025dapo}. We report six mathematical test benchmarks: MATH500~\citep{hendrycks2021math,lightman2023lets}, Minerva~\citep{lewkowycz2022minerva}, AIME24~\citep{aime24}, AIME25~\citep{aime25}, AMC24~\citep{extraordinarylab2025amc}, and OlympiadBench~\citep{he2024olympiadbench}; and three out-of-distribution benchmarks: MMLU-Pro~\citep{wang2024mmlupro}, GPQA-Diamond~\citep{rein2023gpqa}, and BBH~\citep{suzgun2022bbh}.

\paragraph{Baselines.}
We compare with standard GRPO and four representative baselines. DAPO~\citep{yu2025dapo} dynamically identifies and discards zero-variance prompts; Dr.GRPO~\citep{liu2025r1zero} redesigns advantages and loss aggregation to address length bias; GSPO~\citep{zheng2025gspo} performs importance weighting, clipping, and optimization at the sequence level; and RePO~\citep{li2025repo} reuses complete historical trajectories from a replay buffer. We follow method-specific configurations from the corresponding papers or public implementations.

\paragraph{Implementation details.}
All methods use four epochs, a training and actor mini-batch size of 64, and eight rollouts per problem. The actor is optimized with AdamW (learning rate $1\times10^{-6}$, weight decay 0.01), a constant learning-rate schedule, and no warmup; we use neither KL loss nor entropy regularization. Training rollouts use temperature 1.0, top-$p$ 1.0, top-$k=-1$, and maximum prompt/response lengths of 1024/4096, while final evaluation uses temperature 0.6, top-$p$ 0.95, top-$k=0$, and maximum response length 4096. For ASE, we use $K=3$ steering clusters plus a clean route, steering strength $\alpha=0.5$, and EMA coefficient $\beta=0.5$. Training uses verl~\citep{sheng2025hybridflow}, vLLM~\citep{kwon2023pagedattention}, BF16, and FlashAttention-2~\citep{dao2024flashattention2} on four NVIDIA A800 GPUs for the 1.7B model and eight for the 8B model. Math-Verify\footnote{\url{https://github.com/huggingface/Math-Verify}} checks final answers. Shared training and evaluation settings are given in Appendix~\ref{app:implementation}; checkpoint-selection and remaining method-specific details are provided in Appendices~\ref{app:checkpoint} and~\ref{app:implementation}.

\subsection{Main Results}
Tables~\ref{tab:8b} and~\ref{tab:17b} present the main results for Qwen3-8B-Base and Qwen3-1.7B-Base, respectively. Each table cell reports Mean@8 / Pass@8 in percent. Mean@8 is the average accuracy over eight sampled answers; Pass@8 is the fraction of problems with at least one correct answer. Reported scores are averaged over three runs. Math Avg. and OOD Avg. are the unweighted means of benchmark-level scores across the six mathematical and three out-of-distribution benchmarks, respectively, computed separately for Mean@8 and Pass@8.

\begin{table*}[t]
\centering
\setlength{\tabcolsep}{-1pt}
\caption{Main results for Qwen3-8B-Base trained on DAPO-Math-17k. Scores are reported as Mean@8 / Pass@8 (in percent). Reported scores are averaged over three runs. Each $\Delta$ row reports the absolute difference from the corresponding baseline in percentage points.}
\label{tab:8b}
\resizebox{\textwidth}{!}{%
\begin{tabular}{lcccccc>{\columncolor{gray!15}}c ccc >{\columncolor{gray!15}}c}
\toprule
& \multicolumn{7}{c}{Mathematical benchmarks} & \multicolumn{4}{c}{OOD benchmarks} \\
\cmidrule(lr){2-8}\cmidrule(lr){9-12}
Method & MATH500 & Minerva & AIME24 & AIME25 & AMC24 & Olympiad & Math Avg. & MMLU-Pro & GPQA-D & BBH & OOD Avg.\\
\midrule
GRPO & \score{88.17}{94.87} & \score{\textbf{43.98}}{55.76} & \score{26.39}{48.89} & \score{21.39}{\textbf{37.78}} & \score{58.06}{82.22} & \score{55.80}{70.18} & \score{48.96}{64.95} & \score{\textbf{67.62}}{85.64} & \score{49.45}{79.80} & \score{76.73}{91.19} & \score{64.60}{85.54}\\
DAPO & \score{86.09}{93.20} & \score{42.56}{54.17} & \score{21.53}{41.11} & \score{18.06}{30.00} & \score{48.98}{74.07} & \score{51.38}{66.82} & \score{44.77}{59.89} & \score{61.28}{83.38} & \score{46.70}{80.81} & \score{76.71}{91.32} & \score{61.56}{85.17}\\
Dr.GRPO & \score{87.83}{\textbf{95.87}} & \score{43.20}{\textbf{59.44}} & \score{25.42}{47.78} & \score{20.69}{\textbf{37.78}} & \score{55.56}{77.78} & \score{57.78}{\textbf{74.78}} & \score{48.41}{\textbf{65.57}} & \score{65.20}{\textbf{86.00}} & \score{48.34}{\textbf{84.85}} & \score{\textbf{77.79}}{91.99} & \score{63.78}{\textbf{87.61}}\\
GSPO & \score{88.05}{95.00} & \score{43.01}{55.51} & \score{25.00}{\textbf{53.33}} & \score{\textbf{22.50}}{36.67} & \score{58.33}{80.00} & \score{56.71}{72.40} & \score{48.94}{65.49} & \score{64.94}{84.08} & \score{47.47}{80.81} & \score{75.74}{91.20} & \score{62.72}{85.36}\\
RePO & \score{86.09}{94.53} & \score{41.57}{54.29} & \score{22.64}{42.22} & \score{18.61}{32.22} & \score{53.06}{81.48} & \score{53.34}{71.36} & \score{45.89}{62.69} & \score{61.21}{84.40} & \score{43.96}{81.31} & \score{67.92}{91.35} & \score{57.69}{85.69}\\
\midrule
\textbf{ASE-GRPO} & \score{\textbf{89.31}}{95.73} & \score{43.70}{56.74} & \score{\textbf{28.19}}{46.67} & \score{21.67}{36.67} & \score{\textbf{62.78}}{\textbf{82.96}} & \score{\textbf{59.32}}{74.09} & \score{\textbf{50.83}}{65.48} & \score{66.63}{85.68} & \score{\textbf{51.01}}{82.32} & \score{76.26}{\textbf{92.74}} & \score{\textbf{64.63}}{86.91}\\
$\Delta$ vs. GRPO & \score{+1.14}{+0.86} & \score{-0.28}{+0.98} & \score{+1.80}{-2.22} & \score{+0.28}{-1.11} & \score{+4.72}{+0.74} & \score{+3.52}{+3.91} & \score{+1.87}{+0.53} & \score{-0.99}{+0.04} & \score{+1.56}{+2.52} & \score{-0.47}{+1.55} & \score{+0.03}{+1.37}\\
$\Delta$ vs. DAPO & \score{+3.22}{+2.53} & \score{+1.14}{+2.57} & \score{+6.66}{+5.56} & \score{+3.61}{+6.67} & \score{+13.80}{+8.89} & \score{+7.94}{+7.27} & \score{+6.06}{+5.59} & \score{+5.35}{+2.30} & \score{+4.31}{+1.51} & \score{-0.45}{+1.42} & \score{+3.07}{+1.74}\\
$\Delta$ vs. Dr.GRPO & \score{+1.48}{-0.14} & \score{+0.50}{-2.70} & \score{+2.77}{-1.11} & \score{+0.98}{-1.11} & \score{+7.22}{+5.18} & \score{+1.54}{-0.69} & \score{+2.42}{-0.09} & \score{+1.43}{-0.32} & \score{+2.67}{-2.53} & \score{-1.53}{+0.75} & \score{+0.85}{-0.70}\\
$\Delta$ vs. GSPO & \score{+1.26}{+0.73} & \score{+0.69}{+1.23} & \score{+3.19}{-6.66} & \score{-0.83}{+0.00} & \score{+4.45}{+2.96} & \score{+2.61}{+1.69} & \score{+1.89}{-0.01} & \score{+1.69}{+1.60} & \score{+3.54}{+1.51} & \score{+0.52}{+1.54} & \score{+1.91}{+1.55}\\
$\Delta$ vs. RePO & \score{+3.22}{+1.20} & \score{+2.13}{+2.45} & \score{+5.55}{+4.45} & \score{+3.06}{+4.45} & \score{+9.72}{+1.48} & \score{+5.98}{+2.73} & \score{+4.94}{+2.79} & \score{+5.42}{+1.28} & \score{+7.05}{+1.01} & \score{+8.34}{+1.39} & \score{+6.94}{+1.22}\\
\bottomrule
\end{tabular}}
\end{table*}

\begin{table*}[t]
\centering
\setlength{\tabcolsep}{-1pt}
\caption{Main results for Qwen3-1.7B-Base trained on DAPO-Math-17k. Scores are reported as Mean@8 / Pass@8 (in percent). Reported scores are averaged over three runs. Each $\Delta$ row reports the absolute difference from the corresponding baseline in percentage points.}
\label{tab:17b}
\resizebox{\textwidth}{!}{%
\begin{tabular}{lcccccc>{\columncolor{gray!15}}c ccc >{\columncolor{gray!15}}c}
\toprule
& \multicolumn{7}{c}{Mathematical benchmarks} & \multicolumn{4}{c}{OOD benchmarks} \\
\cmidrule(lr){2-8}\cmidrule(lr){9-12}
Method & MATH500 & Minerva & AIME24 & AIME25 & AMC24 & Olympiad & Math Avg. & MMLU-Pro & GPQA-D & BBH & OOD Avg.\\
\midrule
GRPO & \score{65.59}{83.47} & \score{27.37}{43.50} & \score{6.39}{14.44} & \score{5.42}{\textbf{22.22}} & \score{18.24}{40.74} & \score{29.82}{50.59} & \score{25.47}{42.50} & \score{35.46}{69.09} & \score{26.45}{79.12} & \score{44.70}{84.15} & \score{35.54}{77.45}\\
DAPO & \score{68.02}{84.20} & \score{27.94}{45.22} & \score{10.00}{\textbf{26.67}} & \score{5.00}{18.89} & \score{21.20}{43.70} & \score{30.87}{50.49} & \score{27.17}{44.86} & \score{39.22}{69.27} & \score{29.12}{76.60} & \score{50.43}{85.28} & \score{39.59}{77.05}\\
Dr.GRPO & \score{68.16}{84.53} & \score{\textbf{30.39}}{45.59} & \score{\textbf{12.08}}{23.33} & \score{2.92}{12.22} & \score{21.39}{40.74} & \score{31.43}{51.68} & \score{27.73}{43.02} & \score{40.54}{70.33} & \score{\textbf{30.30}}{77.61} & \score{51.99}{85.43} & \score{40.94}{77.79}\\
GSPO & \score{68.42}{84.80} & \score{28.71}{44.36} & \score{9.58}{22.22} & \score{4.86}{18.89} & \score{18.24}{39.26} & \score{32.00}{53.81} & \score{26.97}{43.89} & \score{38.38}{70.15} & \score{27.90}{78.96} & \score{47.02}{84.75} & \score{37.77}{77.95}\\
RePO & \score{65.42}{84.67} & \score{26.19}{44.61} & \score{7.08}{22.22} & \score{4.31}{15.56} & \score{17.78}{42.22} & \score{29.60}{51.88} & \score{25.06}{43.53} & \score{36.15}{69.45} & \score{26.54}{76.26} & \score{44.36}{84.96} & \score{35.68}{76.89}\\
\midrule
\textbf{ASE-GRPO} & \score{\textbf{71.15}}{\textbf{88.40}} & \score{29.78}{\textbf{47.06}} & \score{10.83}{25.56} & \score{\textbf{8.19}}{21.11} & \score{\textbf{27.96}}{\textbf{56.30}} & \score{\textbf{34.12}}{\textbf{54.65}} & \score{\textbf{30.34}}{\textbf{48.85}} & \score{\textbf{41.35}}{\textbf{71.12}} & \score{29.95}{\textbf{79.97}} & \score{\textbf{52.46}}{\textbf{85.87}} & \score{\textbf{41.25}}{\textbf{78.98}}\\
$\Delta$ vs. GRPO & \score{+5.56}{+4.93} & \score{+2.41}{+3.56} & \score{+4.44}{+11.12} & \score{+2.77}{-1.11} & \score{+9.72}{+15.56} & \score{+4.30}{+4.06} & \score{+4.87}{+6.35} & \score{+5.89}{+2.03} & \score{+3.50}{+0.85} & \score{+7.76}{+1.72} & \score{+5.71}{+1.53}\\
$\Delta$ vs. DAPO & \score{+3.13}{+4.20} & \score{+1.84}{+1.84} & \score{+0.83}{-1.11} & \score{+3.19}{+2.22} & \score{+6.76}{+12.60} & \score{+3.25}{+4.16} & \score{+3.17}{+3.99} & \score{+2.13}{+1.85} & \score{+0.83}{+3.37} & \score{+2.03}{+0.59} & \score{+1.66}{+1.93}\\
$\Delta$ vs. Dr.GRPO & \score{+2.99}{+3.87} & \score{-0.61}{+1.47} & \score{-1.25}{+2.23} & \score{+5.27}{+8.89} & \score{+6.57}{+15.56} & \score{+2.69}{+2.97} & \score{+2.61}{+5.83} & \score{+0.81}{+0.79} & \score{-0.35}{+2.36} & \score{+0.47}{+0.44} & \score{+0.31}{+1.19}\\
$\Delta$ vs. GSPO & \score{+2.73}{+3.60} & \score{+1.07}{+2.70} & \score{+1.25}{+3.34} & \score{+3.33}{+2.22} & \score{+9.72}{+17.04} & \score{+2.12}{+0.84} & \score{+3.37}{+4.96} & \score{+2.97}{+0.97} & \score{+2.05}{+1.01} & \score{+5.44}{+1.12} & \score{+3.48}{+1.03}\\
$\Delta$ vs. RePO & \score{+5.73}{+3.73} & \score{+3.59}{+2.45} & \score{+3.75}{+3.34} & \score{+3.88}{+5.55} & \score{+10.18}{+14.08} & \score{+4.52}{+2.77} & \score{+5.28}{+5.32} & \score{+5.20}{+1.67} & \score{+3.41}{+3.71} & \score{+8.10}{+0.91} & \score{+5.57}{+2.09}\\
\bottomrule
\end{tabular}}
\end{table*}

ASE-GRPO achieves the highest mathematical Mean@8 average on both model sizes. On Qwen3-1.7B-Base, it reaches 30.34, compared with 25.47 for GRPO and 27.73 for Dr.GRPO; its mathematical Pass@8 also leads at 48.85. On Qwen3-8B-Base, it reaches 50.83, exceeding GRPO by 1.87 points and GSPO by 1.89 points. Improvements are particularly clear on AMC24 and OlympiadBench, while other methods remain stronger on Minerva or AIME24.

The gains extend beyond the mathematical training domain. ASE-GRPO obtains the highest OOD Mean@8 average on both Qwen3-1.7B-Base (41.25) and Qwen3-8B-Base (64.63), compared with 35.54 and 64.60 for GRPO, respectively. The 1.7B model also achieves the highest OOD Pass@8 average, whereas Dr.GRPO remains higher on this metric for the 8B model. Together, these results suggest that activation-space exploration improves the unsteered policy across model scales, with consistent gains in mathematical reasoning and evidence of transfer to out-of-distribution tasks.

Figure~\ref{fig:pass-at-k} extends the evaluation to larger sampling budgets. ASE-GRPO achieves higher math-benchmark Pass@$k$ than GRPO across $k=8$ to $128$ on both model scales. This suggests that the learned policy is more effective at discovering correct solutions when sampling multiple candidate responses.

\subsection{Dynamics of Training and Vector-Bank}
\begin{figure*}[t]
\centering
\begin{minipage}[t]{0.95in}
\vspace{0pt}
\centering
\includegraphics[width=\linewidth]{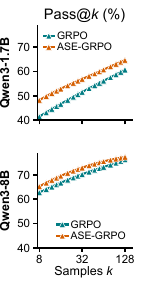}
\captionof{figure}{Pass@$k$ averaged over six math benchmarks (in percent).}
\label{fig:pass-at-k}
\end{minipage}%
\hfill
\begin{minipage}[t]{3.4in}
\vspace{0pt}
\centering
\includegraphics[width=\linewidth]{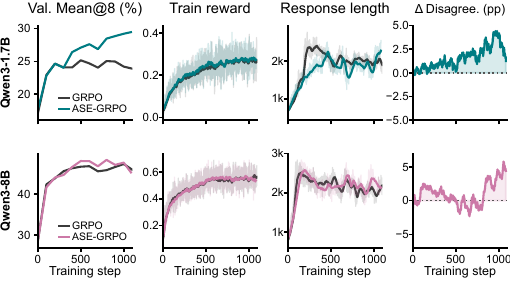}
\captionof{figure}{Training dynamics of GRPO and ASE-GRPO on Qwen3-1.7B-Base and Qwen3-8B-Base. The four columns show validation Mean@8, training reward, response length, and the rollout disagreement gap across training steps.}
\label{fig:training-dynamics}
\end{minipage}%
\hfill
\begin{minipage}[t]{0.9in}
\vspace{0pt}
\centering
\includegraphics[width=\linewidth]{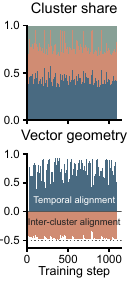}
\captionof{figure}{Vector-bank dynamics for Qwen3-1.7B-Base.}
\label{fig:bank-dynamics}
\end{minipage}
\end{figure*}

Figure~\ref{fig:training-dynamics} tracks validation performance and rollout behavior on Qwen3-1.7B-Base and Qwen3-8B-Base. Validation Mean@8 rises during training, and ASE-GRPO reaches higher values than GRPO on both model scales. Training reward follows a similar trajectory for the two methods, and response lengths remain within a comparable range. We use pairwise disagreement between answers sampled for the same problem as an answer-level measure of exploration. The disagreement gap (ASE-GRPO minus GRPO) is positive overall and generally grows over training on both model sizes, showing that ASE-GRPO explores a broader range of answers alongside its validation gains without a marked change in training reward or response length.

Figure~\ref{fig:bank-dynamics} shows how ASE-GRPO's vector bank evolves on Qwen3-1.7B-Base. Each of the three cluster shares is the fraction of activation pairs assigned to that cluster at a bank update. The shares remain broadly stable and reasonably balanced throughout training, showing that all three directions continue to receive support from the contrast data. In the \emph{Vector geometry} panel, temporal alignment is the mean cosine similarity between newly extracted vectors and their matched vectors from the previous bank; the median cosine of 0.73 shows that successive vectors remain similarly oriented. Inter-cluster alignment is the mean pairwise cosine similarity among directions in the updated bank; values near the three-vector lower bound of $-0.5$ show that the directions remain well separated.

\subsection{Inference-Time Steering versus Training-Time Steering}
Activation steering can be used in two distinct ways. An inference-time intervention changes the output distribution of a fixed model during evaluation, whereas our training-time intervention diversifies the rollouts used to train the policy. We compare these two uses of steering to determine whether the main benefit comes from the direct effect of steering a fixed policy or from cumulative gains of online exploration during training.
We compare six conditions: Base, Base$^{\mathrm{S}}$, GRPO, GRPO$^{\mathrm{S}}$, ASE-GRPO, and ASE-GRPO$^{\mathrm{S}}$, where $^{\mathrm{S}}$ denotes inference-time intervention. To assess the potential gains from inference-time steering, we sweep its parameters separately for each checkpoint; details are reported in Appendix~\ref{app:additional-ab}.

Table~\ref{tab:inference-steering} summarizes the comparison. Inference-time intervention provides only marginal changes for the untrained and GRPO checkpoints: Base$^{\mathrm{S}}$ changes Math Mean@8 from 18.74 to 18.86 and GRPO$^{\mathrm{S}}$ changes it from 25.47 to 25.81. ASE-GRPO already outperforms standard GRPO without inference-time intervention, and applying inference-time intervention to ASE-GRPO gives a further 0.51-point Math Mean@8 increase and 0.15-point OOD Mean@8 increase, while OOD Pass@8 decreases from 78.98 to 78.44. These results indicate that the main gain comes from training-time exploration, with only modest direct effects from steering a fixed policy.

\subsection{Ablating Verifier-Guided Direction}
The vector bank is estimated from within-problem success--failure contrasts. This raises a specific attribution question: does verifier feedback matter, or is the gain explained by injecting any additional activation perturbation? We therefore design a strictly matched ablation in which only the construction of the exploration directions varies. Random directions replace each cluster candidate with an independent random vector of the same norm; shuffled labels randomly swap the positive and negative labels within each pair before clustering and direction estimation; and reversed directions negate the learned vectors while preserving their cluster assignments. These controls retain the same downstream bank update and routing procedure.

\begin{figure*}[t]
\centering
\begin{minipage}[t]{0.465\textwidth}
\vspace{0pt}
\centering
\scriptsize
\captionof{table}{Training- versus inference-time steering on Qwen3-1.7B-Base. $^{\mathrm{S}}$ denotes inference-time steering. Math/OOD Avg. report Mean@8 / Pass@8 (\%), averaged over three evaluations.}
\label{tab:inference-steering}
\resizebox{0.65\linewidth}{!}{%
\begin{tabular}{lcc}
\toprule
Condition & Math Avg. & OOD Avg. \\
\midrule
Base & $18.74/39.31$ & $18.02/61.16$ \\
Base$^{\mathrm{S}}$ & $18.86/40.09$ & $17.93/61.51$ \\
GRPO & $25.47/42.50$ & $35.54/77.45$ \\
GRPO$^{\mathrm{S}}$ & $25.81/43.13$ & $35.83/77.29$ \\
\rowcolor{gray!15}
ASE-GRPO & $30.34/48.85$ & $41.25/78.98$ \\
\rowcolor{gray!15}
ASE-GRPO$^{\mathrm{S}}$ & $30.85/49.09$ & $41.40/78.44$ \\
\bottomrule
\end{tabular}
}
\end{minipage}
\hfill
\begin{minipage}[t]{0.49\textwidth}
\vspace{0pt}
\centering
\scriptsize
\captionof{table}{Direction-construction ablation on Qwen3-1.7B-Base. ASE-GRPO variants differ only in direction construction. Math/OOD Avg. report Mean@8 / Pass@8 (\%), averaged over three evaluations.}
\label{tab:matched-direction}
\resizebox{0.65\linewidth}{!}{%
\begin{tabular}{lcc}
\toprule
Condition & Math Avg. & OOD Avg. \\
\midrule
GRPO & 25.47/42.50 & 35.54/77.45 \\
Random directions & 27.76/45.62 & 39.48/78.09 \\
Shuffled labels & 27.56/44.46 & 39.93/77.03 \\
Reversed directions & 27.87/45.15 & 39.13/78.48 \\
\rowcolor{gray!15}
ASE-GRPO & 30.34/48.85 & 41.25/78.98 \\
\bottomrule
\end{tabular}
}
\end{minipage}
\end{figure*}

As shown in Table~\ref{tab:matched-direction}, ASE-GRPO outperforms all three ablation variants on both aggregates, reaching 30.34/48.85 on Math Avg. and 41.25/78.98 on OOD Avg. The variants mostly exceed GRPO (25.47/42.50 and 35.54/77.45, respectively), except that shuffled labels has lower OOD Pass@8 (77.03 versus 77.45).
These variants retain the clustered $K=3$ bank, routing, and EMA updates, which may contribute to their gains over GRPO. An additional control without clustering or EMA yields only a small increase in Math Mean@8 and a decrease in Pass@8 relative to GRPO (Appendix~\ref{app:matched-direction}), suggesting that arbitrary random steering alone does not provide a consistent benefit. Together, these results suggest that the shared bank structure and EMA may account for part of the improvement, while verifier-guided directions provide additional gains within the matched setup.

\subsection{Evaluation on Alternative Policy Optimization Formulations}
To test whether the benefit of ASE is specific to the standard GRPO formulation, we apply the same training-time exploration mechanism to Dr.GRPO and GSPO. We use Qwen3-1.7B-Base, the same four-epoch training budget, and eight rollouts per problem for all conditions. We do not perform a separate parameter search for either alternative formulation and use the preset ASE configuration. Table~\ref{tab:gspo-ase} summarizes the results. Compared with Dr.GRPO, ASE-Dr.GRPO increases Math Avg. Mean@8 and Pass@8 by 1.28 and 2.18 points, respectively, and OOD Avg. Mean@8 by 0.36 points. Compared with GSPO, ASE-GSPO increases Math Avg. Mean@8 and Pass@8 by 1.49 and 0.27 points, respectively, and OOD Avg. Mean@8 and Pass@8 by 2.08 and 0.34 points, respectively. These results indicate that activation-space exploration also benefits alternative policy-optimization formulations.

\begin{table*}[t]
\centering
\setlength{\tabcolsep}{-1pt}
\caption{ASE-Dr.GRPO and ASE-GSPO on Qwen3-1.7B-Base. Entries are Mean@8 / Pass@8 (\%); Math Avg. and OOD Avg. cover six mathematical and three OOD benchmarks, respectively.}
\label{tab:gspo-ase}
\resizebox{\textwidth}{!}{%
\begin{tabular}{lcccccc>{\columncolor{gray!15}}c ccc >{\columncolor{gray!15}}c}
\toprule
& \multicolumn{7}{c}{Mathematical benchmarks} & \multicolumn{4}{c}{OOD benchmarks} \\
\cmidrule(lr){2-8}\cmidrule(lr){9-12}
Method & MATH500 & Minerva & AIME24 & AIME25 & AMC24 & Olympiad & Math Avg. & MMLU-Pro & GPQA-D & BBH & OOD Avg. \\
\midrule
Dr.GRPO & \score{68.27}{84.60} & \score{\textbf{31.25}}{47.06} & \score{\textbf{12.08}}{20.00} & \score{2.92}{\textbf{16.67}} & \score{20.83}{40.00} & \score{31.36}{\textbf{52.67}} & \score{27.79}{43.50} & \score{40.36}{\textbf{70.62}} & \score{29.86}{77.27} & \score{\textbf{52.03}}{\textbf{85.61}} & \score{40.75}{\textbf{77.83}} \\
ASE-Dr.GRPO & \score{\textbf{69.40}}{\textbf{86.20}} & \score{30.06}{\textbf{47.43}} & \score{11.25}{\textbf{26.67}} & \score{\textbf{5.83}}{\textbf{16.67}} & \score{\textbf{24.17}}{\textbf{44.44}} & \score{\textbf{33.70}}{\textbf{52.67}} & \score{\textbf{29.07}}{\textbf{45.68}} & \score{\textbf{40.45}}{69.23} & \score{\textbf{31.06}}{\textbf{77.78}} & \score{51.81}{84.95} & \score{\textbf{41.11}}{77.32} \\
$\Delta$ vs. Dr.GRPO & \score{+1.13}{+1.60} & \score{-1.19}{+0.37} & \score{-0.83}{+6.67} & \score{+2.91}{+0.00} & \score{+3.34}{+4.44} & \score{+2.34}{+0.00} & \score{+1.28}{+2.18} & \score{+0.09}{-1.39} & \score{+1.20}{+0.51} & \score{-0.22}{-0.66} & \score{+0.36}{-0.51} \\
\midrule
GSPO & \score{69.00}{86.20} & \score{28.72}{43.75} & \score{8.75}{\textbf{23.33}} & \score{5.00}{\textbf{23.33}} & \score{19.17}{42.22} & \score{\textbf{32.64}}{54.30} & \score{27.21}{45.52} & \score{38.15}{70.48} & \score{26.96}{\textbf{80.81}} & \score{46.91}{84.55} & \score{37.34}{78.61} \\
ASE-GSPO & \score{\textbf{69.80}}{\textbf{87.00}} & \score{\textbf{30.19}}{\textbf{47.43}} & \score{\textbf{10.00}}{\textbf{23.33}} & \score{\textbf{6.25}}{13.33} & \score{\textbf{23.33}}{\textbf{48.89}} & \score{32.60}{\textbf{54.75}} & \score{\textbf{28.70}}{\textbf{45.79}} & \score{\textbf{40.48}}{\textbf{72.52}} & \score{\textbf{28.98}}{78.79} & \score{\textbf{48.80}}{\textbf{85.53}} & \score{\textbf{39.42}}{\textbf{78.95}} \\
$\Delta$ vs. GSPO & \score{+0.80}{+0.80} & \score{+1.47}{+3.68} & \score{+1.25}{+0.00} & \score{+1.25}{-10.00} & \score{+4.16}{+6.67} & \score{-0.04}{+0.45} & \score{+1.49}{+0.27} & \score{+2.33}{+2.04} & \score{+2.02}{-2.02} & \score{+1.89}{+0.98} & \score{+2.08}{+0.34} \\
\bottomrule
\end{tabular}}
\end{table*}

\subsection{Training-time overhead.}
Table~\ref{tab:training-overhead} compares latency and throughput for the matched GRPO and ASE-GRPO runs, averaged over the training trajectory after excluding the first 10 warm-up steps. ASE-GRPO increases token-normalized rollout latency by only 2.72\% on the 1.7B model and 2.51\% on the 8B model, while end-to-end throughput decreases by 6.29\% and 6.44\%, respectively. The additional activation replay and vector-bank maintenance are reflected in the actor-update latency, but the resulting mean step-time increases remain modest at 0.91\% and 7.19\%. Moreover, the vector bank is discarded after training, so deployment incurs no additional inference latency.

\begin{table*}[t]
\centering
\scriptsize
\setlength{\tabcolsep}{2pt}
\caption{Training-time overhead on matched GRPO and ASE-GRPO runs. Throughput uses all prompt and response tokens, rollout latency is normalized by output tokens, and actor-update latency is normalized by all valid tokens.}
\label{tab:training-overhead}
\resizebox{0.60\textwidth}{!}{%
\begin{tabular}{llcccc}
\toprule
Model & Method & \shortstack{Step time\\(s/step) $\downarrow$} & \shortstack{Training throughput\\(k tokens/s/GPU) $\uparrow$} & \shortstack{Rollout latency\\($\mu$s/output token) $\downarrow$} & \shortstack{Actor-update time\\($\mu$s/valid token) $\downarrow$} \\
\midrule
1.7B & GRPO & 92.21 & 2.845 & 52.91 & 29.15 \\
1.7B & ASE-GRPO & 93.05 & 2.666 & 54.35 & 33.60 \\
1.7B & $\Delta$ vs. GRPO & +0.91\% & $-$6.29\% & +2.72\% & +15.27\% \\
\midrule
8B & GRPO & 136.10 & 1.041 & 60.08 & 49.35 \\
8B & ASE-GRPO & 145.88 & 0.974 & 61.59 & 56.08 \\
8B & $\Delta$ vs. GRPO & +7.19\% & $-$6.44\% & +2.51\% & +13.64\% \\
\bottomrule
\end{tabular}
}
\end{table*}

\section{Conclusion}

We introduced Activation-Space Exploration (ASE), an online method that uses verifier feedback to construct and update multiple activation-space directions for training rollouts. By distributing a fixed rollout budget across these directions, ASE improves mathematical reasoning on two model scales and also benefits Dr.GRPO and GSPO. Matched ablations show that success--failure contrasts yield stronger gains than random, label-shuffled, or reversed directions, while comparisons with inference-time steering support the value of using these interventions during training. All steering is removed at evaluation, so the gains persist in the learned base policy without additional inference-time intervention. These findings support activation space as a complementary exploration axis for RLVR.

\subsection*{AI use statement}
In this work, we used generative AI tools to design and provide feedback on research methodology and experiments, assist with the execution of experiments, and assist with translation of parts of the manuscript. We also used generative AI tools for sourcing and searching for research information, identifying relevant literature, creating and editing software code used in the experiments, and formatting references. We did not use generative AI tools to generate synthetic datasets, develop theoretical models or conceptual frameworks, formulate mathematical claims, provide critical ingredients for mathematical proofs, assist in writing proofs, propose or refine hypotheses, clean or reformat datasets, support qualitative or thematic data analysis, or interpret the reported results. All AI-assisted work was reviewed by the authors. Experimental procedures, software code, outputs, citations, translations, and reported claims were checked and verified by the authors. We take responsibility for the final content of this work, including all text, claims, results, and artifacts produced with the aid of generative AI.

\bibliography{iclr2027_conference}

@book{sutton2018reinforcement,
  title     = {Reinforcement Learning: An Introduction},
  author    = {Sutton, Richard S. and Barto, Andrew G.},
  edition   = {2},
  publisher = {MIT Press},
  year      = {2018},
  url       = {http://incompleteideas.net/book/the-book-2nd.html}
}

@inproceedings{le2022coderl,
  title     = {{CodeRL}: Mastering Code Generation Through Pretrained Models and Deep Reinforcement Learning},
  author    = {Le, Hung and Wang, Yue and Gotmare, Akhilesh Deepak and Savarese, Silvio and Hoi, Steven Chu Hong},
  booktitle = {Advances in Neural Information Processing Systems},
  volume    = {35},
  pages     = {21314--21328},
  year      = {2022},
  doi       = {10.52202/068431-1549},
  url       = {https://doi.org/10.52202/068431-1549}
}

@misc{aime24,
  title  = {American Invitational Mathematics Examination ({AIME}) 2024},
  author = {Zhang, Yifan and {Math-AI Team}},
  year   = {2024},
  url    = {https://huggingface.co/datasets/math-ai/aime24}
}

@misc{aime25,
  title  = {American Invitational Mathematics Examination ({AIME}) 2025},
  author = {Zhang, Yifan and {Math-AI Team}},
  year   = {2025},
  url    = {https://huggingface.co/datasets/math-ai/aime25}
}

@misc{extraordinarylab2025amc,
  title  = {{AMC} Dataset},
  author = {{ExtraordinaryLab}},
  year   = {2025},
  url    = {https://huggingface.co/datasets/extraordinarylab/amc}
}

@article{beaglehole2026universal,
  title         = {Toward Universal Steering and Monitoring of {AI} Models},
  author        = {Daniel Beaglehole and Adityanarayanan Radhakrishnan and Enric Boix-Adser{\`a} and Mikhail Belkin},
  journal       = {Science},
  volume        = {391},
  number        = {6787},
  pages         = {787--792},
  year          = {2026},
  doi           = {10.1126/science.aea6792},
  url           = {https://doi.org/10.1126/science.aea6792}
}

@article{chen2025persona,
  title   = {Persona Vectors: Monitoring and Controlling Character Traits in Language Models},
  author  = {Chen, Runjin and Arditi, Andy and Sleight, Henry and Evans, Owain and Lindsey, Jack},
  journal = {arXiv preprint arXiv:2507.21509},
  year    = {2025},
  url     = {https://arxiv.org/abs/2507.21509}
}

@inproceedings{hendrycks2021math,
  title         = {Measuring Mathematical Problem Solving With the {MATH} Dataset},
  author        = {Dan Hendrycks and Collin Burns and Saurav Kadavath and Akul Arora and Steven Basart and Eric Tang and Dawn Song and Jacob Steinhardt},
  year          = {2021},
  booktitle     = {Proceedings of the Neural Information Processing Systems Track on Datasets and Benchmarks},
  volume        = {1},
  url           = {https://datasets-benchmarks-proceedings.neurips.cc/paper/2021/hash/be83ab3ecd0db773eb2dc1b0a17836a1-Abstract-round2.html}
}

@article{lightman2023lets,
  title   = {Let's Verify Step by Step},
  author  = {Hunter Lightman and Vineet Kosaraju and Yura Burda and Harri Edwards and Bowen Baker and Teddy Lee and Jan Leike and John Schulman and Ilya Sutskever and Karl Cobbe},
  journal = {arXiv preprint arXiv:2305.20050},
  year    = {2023},
  url     = {https://arxiv.org/abs/2305.20050}
}

@inproceedings{kwon2023pagedattention,
  title         = {Efficient Memory Management for Large Language Model Serving with {PagedAttention}},
  author        = {Woosuk Kwon and Zhuohan Li and Siyuan Zhuang and Ying Sheng and Lianmin Zheng and Cody Hao Yu and Joseph E. Gonzalez and Hao Zhang and Ion Stoica},
  booktitle     = {Proceedings of the 29th Symposium on Operating Systems Principles},
  pages         = {611--626},
  year          = {2023},
  doi           = {10.1145/3600006.3613165},
  url           = {https://doi.org/10.1145/3600006.3613165}
}

@inproceedings{lewkowycz2022minerva,
  title         = {Solving Quantitative Reasoning Problems with Language Models},
  author        = {Aitor Lewkowycz and Anders Andreassen and David Dohan and Ethan Dyer and Henryk Michalewski and Vinay Ramasesh and Ambrose Slone and Cem Anil and Imanol Schlag and Theo Gutman-Solo and Yuhuai Wu and Behnam Neyshabur and Guy Gur-Ari and Vedant Misra},
  year          = {2022},
  booktitle     = {Advances in Neural Information Processing Systems},
  volume        = {35},
  pages         = {3843--3857},
  doi           = {10.52202/068431-0278},
  url           = {https://doi.org/10.52202/068431-0278}
}

@inproceedings{he2024olympiadbench,
  title         = {{OlympiadBench}: A Challenging Benchmark for Promoting {AGI} with Olympiad-Level Bilingual Multimodal Scientific Problems},
  author        = {Chaoqun He and Renjie Luo and Yuzhuo Bai and Shengding Hu and Zhen Thai and Junhao Shen and Jinyi Hu and Xu Han and Yujie Huang and Yuxiang Zhang and Jie Liu and Lei Qi and Zhiyuan Liu and Maosong Sun},
  year          = {2024},
  booktitle     = {Proceedings of the 62nd Annual Meeting of the Association for Computational Linguistics (Volume 1: Long Papers)},
  pages         = {3828--3850},
  doi           = {10.18653/v1/2024.acl-long.211},
  url           = {https://doi.org/10.18653/v1/2024.acl-long.211}
}

@inproceedings{wang2024mmlupro,
  title         = {{MMLU-Pro}: A More Robust and Challenging Multi-Task Language Understanding Benchmark},
  author        = {Yubo Wang and Xueguang Ma and Ge Zhang and Yuansheng Ni and Abhranil Chandra and Shiguang Guo and Weiming Ren and Aaran Arulraj and Xuan He and Ziyan Jiang and Tianle Li and Max Ku and Kai Wang and Alex Zhuang and Rongqi Fan and Xiang Yue and Wenhu Chen},
  year          = {2024},
  booktitle     = {Advances in Neural Information Processing Systems},
  volume        = {37},
  pages         = {95266--95290},
  doi           = {10.52202/079017-3018},
  url           = {https://doi.org/10.52202/079017-3018}
}

@inproceedings{rein2023gpqa,
  title         = {{GPQA}: A Graduate-Level Google-Proof {Q\&A} Benchmark},
  author        = {David Rein and Betty Li Hou and Asa Cooper Stickland and Jackson Petty and Richard Yuanzhe Pang and Julien Dirani and Julian Michael and Samuel R. Bowman},
  year          = {2024},
  booktitle     = {First Conference on Language Modeling},
  url           = {https://openreview.net/forum?id=Ti67584b98}
}

@inproceedings{suzgun2022bbh,
  title         = {Challenging {BIG-Bench} Tasks and Whether Chain-of-Thought Can Solve Them},
  author        = {Mirac Suzgun and Nathan Scales and Nathanael Schärli and Sebastian Gehrmann and Yi Tay and Hyung Won Chung and Aakanksha Chowdhery and Quoc Le and Ed Chi and Denny Zhou and Jason Wei},
  year          = {2023},
  booktitle     = {Findings of the Association for Computational Linguistics: ACL 2023},
  pages         = {13003--13051},
  doi           = {10.18653/v1/2023.findings-acl.824},
  url           = {https://doi.org/10.18653/v1/2023.findings-acl.824}
}

@misc{cui2025entropy,
  title         = {The Entropy Mechanism of Reinforcement Learning for Reasoning Language Models},
  author        = {Ganqu Cui and Yuchen Zhang and Jiacheng Chen and Lifan Yuan and Zhi Wang and Yuxin Zuo and Haozhan Li and Yuchen Fan and Huayu Chen and Weize Chen and Zhiyuan Liu and Hao Peng and Lei Bai and Wanli Ouyang and Yu Cheng and Bowen Zhou and Ning Ding},
  year          = {2025},
  eprint        = {2505.22617},
  archivePrefix = {arXiv},
  primaryClass  = {cs.LG},
  doi           = {10.48550/arXiv.2505.22617},
  url           = {https://arxiv.org/abs/2505.22617}
}

@misc{cui2025painless,
  title         = {Painless Activation Steering: An Automated, Lightweight Approach for Post-Training Large Language Models},
  author        = {Sasha Cui and Zhongren Chen},
  year          = {2026},
  eprint        = {2509.22739},
  archivePrefix = {arXiv},
  primaryClass  = {cs.CL},
  doi           = {10.48550/arXiv.2509.22739},
  url           = {https://arxiv.org/abs/2509.22739}
}

@inproceedings{liu2025r1zero,
  title         = {Understanding {R1-Zero}-Like Training: A Critical Perspective},
  author        = {Zichen Liu and Changyu Chen and Wenjun Li and Penghui Qi and Tianyu Pang and Chao Du and Wee Sun Lee and Min Lin},
  year          = {2025},
  booktitle     = {Second Conference on Language Modeling},
  url           = {https://openreview.net/forum?id=5PAF7PAY2Y}
}

@misc{zheng2025gspo,
  title         = {Group Sequence Policy Optimization},
  author        = {Chujie Zheng and Shixuan Liu and Mingze Li and Xiong-Hui Chen and Bowen Yu and Chang Gao and Kai Dang and Yuqiong Liu and Rui Men and An Yang and Jingren Zhou and Junyang Lin},
  year          = {2025},
  eprint        = {2507.18071},
  archivePrefix = {arXiv},
  primaryClass  = {cs.LG},
  url           = {https://arxiv.org/abs/2507.18071}
}

@misc{hu2026extra,
  title         = {{ExTra}: Exploratory Trajectory Optimization for Language Model Reinforcement Learning},
  author        = {Wenyang Hu and Junxiang Jia and Zhen Shu and Daniel Dahlmeier and See-Kiong Ng and Bryan Kian Hsiang Low},
  year          = {2026},
  eprint        = {2606.24994},
  archivePrefix = {arXiv},
  primaryClass  = {cs.LG},
  doi           = {10.48550/arXiv.2606.24994},
  url           = {https://arxiv.org/abs/2606.24994}
}

@inproceedings{li2023iti,
  title         = {Inference-Time Intervention: Eliciting Truthful Answers from a Language Model},
  author        = {Kenneth Li and Oam Patel and Fernanda Vi{\'e}gas and Hanspeter Pfister and Martin Wattenberg},
  booktitle     = {Advances in Neural Information Processing Systems},
  volume        = {36},
  pages         = {41451--41530},
  publisher     = {Curran Associates, Inc.},
  year          = {2023},
  doi           = {10.52202/075280-1797},
  url           = {https://doi.org/10.52202/075280-1797}
}

@misc{li2025repo,
  title         = {{RePO}: Replay-Enhanced Policy Optimization},
  author        = {Siheng Li and Zhanhui Zhou and Wai Lam and Chao Yang and Chaochao Lu},
  year          = {2025},
  eprint        = {2506.09340},
  archivePrefix = {arXiv},
  primaryClass  = {cs.CL},
  doi           = {10.48550/arXiv.2506.09340},
  url           = {https://arxiv.org/abs/2506.09340}
}

@article{li2026edo,
  title         = {Exploration-Driven Optimization for Test-Time Large Language Model Reasoning},
  author        = {Changhao Li and Yuchen Zhuang and Chenxiao Gao and Haotian Sun and Rushi Qiang and Chao Zhang and Bo Dai},
  journal       = {Transactions on Machine Learning Research},
  year          = {2026},
  month         = may,
  url           = {https://openreview.net/forum?id=NiINDlzvNj}
}

@inproceedings{rimsky2024steering,
  title         = {Steering Llama 2 via Contrastive Activation Addition},
  author        = {Nina Rimsky and Nick Gabrieli and Julian Schulz and Meg Tong and Evan Hubinger and Alexander Turner},
  booktitle     = {Proceedings of the 62nd Annual Meeting of the Association for Computational Linguistics (Volume 1: Long Papers)},
  pages         = {15504--15522},
  address       = {Bangkok, Thailand},
  publisher     = {Association for Computational Linguistics},
  year          = {2024},
  month         = aug,
  doi           = {10.18653/v1/2024.acl-long.828},
  url           = {https://doi.org/10.18653/v1/2024.acl-long.828}
}

@misc{shao2024deepseekmath,
  title         = {{DeepSeekMath}: Pushing the Limits of Mathematical Reasoning in Open Language Models},
  author        = {Zhihong Shao and Peiyi Wang and Qihao Zhu and Runxin Xu and Junxiao Song and Xiao Bi and Haowei Zhang and Mingchuan Zhang and Y. K. Li and Y. Wu and Daya Guo},
  year          = {2024},
  eprint        = {2402.03300},
  archivePrefix = {arXiv},
  primaryClass  = {cs.CL},
  doi           = {10.48550/arXiv.2402.03300},
  url           = {https://arxiv.org/abs/2402.03300}
}

@misc{qwen3technicalreport,
  title         = {{Qwen3} Technical Report},
  author        = {{Qwen Team}},
  year          = {2025},
  eprint        = {2505.09388},
  archivePrefix = {arXiv},
  primaryClass  = {cs.CL},
  url           = {https://arxiv.org/abs/2505.09388}
}

@inproceedings{sheng2025hybridflow,
  title         = {{HybridFlow}: A Flexible and Efficient {RLHF} Framework},
  author        = {Guangming Sheng and Chi Zhang and Zilingfeng Ye and Xibin Wu and Wang Zhang and Ru Zhang and Yanghua Peng and Haibin Lin and Chuan Wu},
  booktitle     = {Proceedings of the Twentieth European Conference on Computer Systems},
  pages         = {1279--1297},
  year          = {2025},
  doi           = {10.1145/3689031.3696075},
  url           = {https://doi.org/10.1145/3689031.3696075}
}

@misc{sinii2025smallvectors,
  title         = {Small Vectors, Big Effects: A Mechanistic Study of {RL}-Induced Reasoning via Steering Vectors},
  author        = {Viacheslav Sinii and Nikita Balagansky and Gleb Gerasimov and Daniil Laptev and Yaroslav Aksenov and Vadim Kurochkin and Alexey Gorbatovski and Boris Shaposhnikov and Daniil Gavrilov},
  year          = {2026},
  eprint        = {2509.06608},
  archivePrefix = {arXiv},
  primaryClass  = {cs.LG},
  doi           = {10.48550/arXiv.2509.06608},
  url           = {https://arxiv.org/abs/2509.06608}
}

@inproceedings{stolfo2025instruction,
  title         = {Improving Instruction-Following in Language Models through Activation Steering},
  author        = {Alessandro Stolfo and Vidhisha Balachandran and Safoora Yousefi and Eric Horvitz and Besmira Nushi},
  booktitle     = {International Conference on Learning Representations},
  pages         = {55790--55823},
  year          = {2025},
  url           = {https://proceedings.iclr.cc/paper_files/paper/2025/hash/8c3262a4c965ba9888f120d4f9e13478-Abstract-Conference.html}
}

@misc{turner2023actadd,
  title         = {Steering Language Models With Activation Engineering},
  author        = {Alexander Matt Turner and Lisa Thiergart and Gavin Leech and David Udell and Juan J. Vazquez and Ulisse Mini and Monte MacDiarmid},
  year          = {2023},
  eprint        = {2308.10248},
  archivePrefix = {arXiv},
  primaryClass  = {cs.CL},
  doi           = {10.48550/arXiv.2308.10248},
  url           = {https://arxiv.org/abs/2308.10248}
}

@inproceedings{venhoff2025reasoning,
  title         = {Understanding Reasoning in Thinking Language Models via Steering Vectors},
  author        = {Constantin Venhoff and Iv{\'a}n Arcuschin and Philip Torr and Arthur Conmy and Neel Nanda},
  booktitle     = {{ICLR} 2025 Workshop on Reasoning and Planning for Large Language Models},
  year          = {2025},
  eprint        = {2506.18167},
  archivePrefix = {arXiv},
  primaryClass  = {cs.LG},
  url           = {https://iclr.cc/virtual/2025/32763}
}

@inproceedings{wang2025highentropy,
  title         = {Beyond the 80/20 Rule: High-Entropy Minority Tokens Drive Effective Reinforcement Learning for {LLM} Reasoning},
  author        = {Shenzhi Wang and Le Yu and Chang Gao and Chujie Zheng and Shixuan Liu and Rui Lu and Kai Dang and Xionghui Chen and Jianxin Yang and Zhenru Zhang and Yuqiong Liu and An Yang and Andrew Zhao and Yang Yue and Shiji Song and Bowen Yu and Gao Huang and Junyang Lin},
  booktitle     = {Advances in Neural Information Processing Systems},
  volume        = {38},
  pages         = {115452--115486},
  publisher     = {Curran Associates, Inc.},
  year          = {2025},
  doi           = {10.52202/085713-3850},
  url           = {https://doi.org/10.52202/085713-3850}
}

@inproceedings{yu2025dapo,
  title         = {{DAPO}: An Open-Source {LLM} Reinforcement Learning System at Scale},
  author        = {Qiying Yu and Zheng Zhang and Ruofei Zhu and Yufeng Yuan and Xiaochen Zuo and Yu Yue and Weinan Dai and Tiantian Fan and Gaohong Liu and Juncai Liu and Lingjun Liu and Xin Liu and Haibin Lin and Zhiqi Lin and Bole Ma and Guangming Sheng and Yuxuan Tong and Chi Zhang and Mofan Zhang and Ru Zhang and Wang Zhang and Hang Zhu and Jinhua Zhu and Jiaze Chen and Jiangjie Chen and Chengyi Wang and Hongli Yu and Yuxuan Song and Xiangpeng Wei and Hao Zhou and Jingjing Liu and Wei-Ying Ma and Ya-Qin Zhang and Lin Yan and Yonghui Wu and Mingxuan Wang},
  booktitle     = {Advances in Neural Information Processing Systems},
  volume        = {38},
  pages         = {113222--113244},
  publisher     = {Curran Associates, Inc.},
  year          = {2025},
  doi           = {10.52202/085713-3775},
  url           = {https://doi.org/10.52202/085713-3775}
}

@inproceedings{yue2025rlvr,
  title         = {Does Reinforcement Learning Really Incentivize Reasoning Capacity in {LLM}s Beyond the Base Model?},
  author        = {Yang Yue and Zhiqi Chen and Rui Lu and Andrew Zhao and Zhaokai Wang and Yang Yue and Shiji Song and Gao Huang},
  booktitle     = {Advances in Neural Information Processing Systems},
  volume        = {38},
  pages         = {57654--57689},
  publisher     = {Curran Associates, Inc.},
  year          = {2025},
  doi           = {10.52202/085713-1933},
  url           = {https://doi.org/10.52202/085713-1933}
}

@inproceedings{zheng2025greso,
  title         = {Act Only When It Pays: Efficient Reinforcement Learning for {LLM} Reasoning via Selective Rollouts},
  author        = {Haizhong Zheng and Yang Zhou and Brian Bartoldson and Bhavya Kailkhura and Fan Lai and Jiawei Zhao and Beidi Chen},
  booktitle     = {Advances in Neural Information Processing Systems},
  volume        = {38},
  pages         = {124321--124346},
  publisher     = {Curran Associates, Inc.},
  year          = {2025},
  doi           = {10.52202/085713-4143},
  url           = {https://doi.org/10.52202/085713-4143}
}

@inproceedings{liao2025enhancing,
  title         = {Enhancing Efficiency and Exploration in Reinforcement Learning for {LLM}s},
  author        = {Mengqi Liao and Xiangyu Xi and Ruinian Chen and Jia Leng and Yangen Hu and Ke Zeng and Shuai Liu and Huaiyu Wan},
  booktitle     = {Proceedings of the 2025 Conference on Empirical Methods in Natural Language Processing},
  pages         = {1451--1463},
  address       = {Suzhou, China},
  publisher     = {Association for Computational Linguistics},
  year          = {2025},
  doi           = {10.18653/v1/2025.emnlp-main.75},
  url           = {https://aclanthology.org/2025.emnlp-main.75/}
}

@inproceedings{dang2026temperature,
  title         = {Temperature as a Meta-Policy: Adaptive Temperature in {LLM} Reinforcement Learning},
  author        = {Haoran Dang and Cuiling Lan and Hai Wan and Xibin Zhao and Yan Lu},
  booktitle     = {International Conference on Learning Representations},
  year          = {2026},
  url           = {https://openreview.net/forum?id=AoTHU2OmS6}
}

@misc{chen2025selfevolving,
  title         = {Self-Evolving Curriculum for {LLM} Reasoning},
  author        = {Xiaoyin Chen and Jiarui Lu and Minsu Kim and Dinghuai Zhang and Jian Tang and Alexandre Pich{\'e} and Nicolas Gontier and Yoshua Bengio and Ehsan Kamalloo},
  year          = {2025},
  eprint        = {2505.14970},
  archivePrefix = {arXiv},
  url           = {https://arxiv.org/abs/2505.14970}
}

@misc{zou2023repe,
  title         = {Representation Engineering: A Top-Down Approach to {AI} Transparency},
  author        = {Andy Zou and Long Phan and Sarah Chen and James Campbell and Phillip Guo and Richard Ren and Alexander Pan and Xuwang Yin and Mantas Mazeika and Ann-Kathrin Dombrowski and Shashwat Goel and Nathaniel Li and Michael J. Byun and Zifan Wang and Alex Mallen and Steven Basart and Sanmi Koyejo and Dawn Song and Matt Fredrikson and J. Zico Kolter and Dan Hendrycks},
  year          = {2023},
  eprint        = {2310.01405},
  archivePrefix = {arXiv},
  primaryClass  = {cs.LG},
  doi           = {10.48550/arXiv.2310.01405},
  url           = {https://arxiv.org/abs/2310.01405}
}

@inproceedings{dao2024flashattention2,
  title     = {Flash{A}ttention-2: Faster Attention with Better Parallelism and Work Partitioning},
  author    = {Dao, Tri},
  booktitle = {International Conference on Learning Representations},
  year      = {2024}
}
\bibliographystyle{iclr2027_conference}

\appendix
\section{Complete Training Algorithm}
\label{app:algorithm}
Algorithm~\ref{alg:ase} summarizes the implementation of ASE-GRPO.
\begin{algorithm}[h]
\caption{ASE-GRPO.}
\label{alg:ase}
\small
\begin{minipage}{0.98\linewidth}
\setlength{\parindent}{0pt}
\setlength{\parskip}{2pt}
\textbf{Input:} initial policy $\pi_{\theta^{(0)}}$, training distribution $\mathcal D$, rollout count $N$, bank size $K$, steering strength $\alpha$, EMA coefficient $\beta$, minimum pair count $B_{\min}$, and update interval $\Delta$.

\textbf{Initialize:} $\mathcal V^{(0)}\leftarrow\emptyset$ and paired-activation buffer $\mathcal B\leftarrow\emptyset$.

\begin{enumerate}[leftmargin=1.7em,label=\arabic*.,itemsep=2pt,topsep=2pt,parsep=0pt]
    \item For $t=0,1,2,\ldots$, sample a batch of problems from $\mathcal D$ and generate routed rollouts with $\pi_{\theta^{(t)}}$ and $\mathcal V^{(t)}$.
    \item Compute binary verifier rewards, then balance and pair correct and incorrect trajectories within each group.
    \item Replay the selected trajectories with the unsteered actor, read the corresponding activations, and append the paired activations to $\mathcal B$.
    \item If $|\mathcal B|\ge B_{\min}$ and at least $\Delta$ iterations have elapsed, fit PCA on the pooled activations and run K-means on the PCA-space pair differences. Compute cluster means in the original hidden space, match candidates to historical directions, apply EMA and normalization, stage $\mathcal V^{(t+1)}$, and clear $\mathcal B$.
    \item Compute group-relative advantages and update the base policy with the clipped GRPO objective.
    \item Synchronize actor weights to the rollout engine. Activate a staged vector bank if one exists; otherwise set $\mathcal V^{(t+1)}\leftarrow\mathcal V^{(t)}$.
\end{enumerate}

\textbf{Output:} the base policy; discard the vector bank.
\end{minipage}
\end{algorithm}

\section{Evaluation and Checkpoint Selection}
\label{app:checkpoint}
Training uses only the local DAPO-Math-17k training file. The original JSONL contains 17,398 samples; filtering by maximum prompt length retains 17,390 samples in the standard configuration, while RePO uses 17,344 samples under its fixed rollout configuration.

We use MATH500 for test-set adaptive model selection by evaluating it periodically during training. For ASE and all baselines, we select the checkpoint with the highest MATH500 Mean@8. We evaluate MATH500 every 100 optimization steps using eight responses per problem, temperature 0.6, top-$p$ 0.95, top-$k=0$, and seed 42. Evaluation on Minerva, AIME24, AIME25, AMC24, OlympiadBench, and the out-of-distribution benchmarks (MMLU-Pro, GPQA-Diamond, and Big-Bench Hard) is performed only on the selected checkpoint for each experiment.

\section{Implementation Details}
\label{app:implementation}
\subsection{Common training and evaluation settings}
\begin{table}[h]
\centering\small
\caption{Common settings for the two model sizes.}
\begin{tabular}{lcc}
\toprule
Configuration & Qwen3-1.7B-Base & Qwen3-8B-Base\\\midrule
Training data & DAPO-Math-17k & DAPO-Math-17k\\
Training epochs & 4 & 4\\
Train/actor mini-batch & 64/64 & 64/64\\
Rollouts per problem & 8 & 8\\
Max prompt/response length & 1024/4096 & 1024/4096\\
Optimizer & AdamW & AdamW\\
Learning rate / weight decay & $1\times10^{-6}$/0.01 & $1\times10^{-6}$/0.01\\
LR scheduler / warmup & constant / 0 & constant / 0\\
Train sampling & 1.0 / 1.0 / $-1$ & 1.0 / 1.0 / $-1$\\
Validation/test sampling & 0.6 / 0.95 / 0 & 0.6 / 0.95 / 0\\
Thinking mode & enabled & enabled\\
Training hardware & 4 NVIDIA A800 GPUs & 8 NVIDIA A800 GPUs\\
\bottomrule
\end{tabular}
\end{table}
All training uses verl~\citep{sheng2025hybridflow}, vLLM~\citep{kwon2023pagedattention} rollouts, BF16, and FlashAttention-2~\citep{dao2024flashattention2}. We filter samples exceeding the maximum prompt length. Except for method-specific objectives or reward processing, we use neither KL loss nor entropy regularization.

\subsection{Baseline and ASE configurations}
\begin{table}[h]
\centering\small
\caption{Method-specific configurations.}
\resizebox{\linewidth}{!}{%
\begin{tabular}{llll}
\toprule
Method & PPO clipping & Loss aggregation / policy loss & Extra processing\\\midrule
GRPO & (0.2, 0.2) & token-mean / vanilla GRPO & none\\
DAPO & (0.2, 0.28) & token-mean / vanilla GRPO & dynamic sampling; overlong buffer\\
Dr.GRPO & (0.2, 0.2) & seq-mean-token-sum-norm / vanilla GRPO & response normalization\\
GSPO & ($3\times10^{-4}$, $4\times10^{-4}$) & seq-mean-token-mean / GSPO & sequence-level loss\\
RePO & (0.2, 0.2) & seq-mean-token-mean / vanilla GRPO & replay cache 16; samples 8\\
ASE-GRPO & (0.2, 0.2) & token-mean / vanilla GRPO & routed activation steering\\
\bottomrule
\end{tabular}}
\end{table}

\begin{table}[h]
\centering\small
\caption{ASE-GRPO configuration.}
\begin{tabular}{ll}
\toprule
Configuration & Value\\\midrule
Intervention layer & Transformer block 0\\
Vector construction & CAA-style clustered positive-minus-negative differences\\
Number of routes $K$ & 3\\
Route allocation & balanced uniform\\
Clean response bank & enabled\\
Readout position function $p$ & response token at 25\% of response length\\
Steering strength $\alpha$ & 0.5\\
EMA coefficient $\beta$ & 0.5\\
Paired buffer minimum $B_{\min}$ & 512 trajectory pairs\\
Vector-bank update interval $\Delta$ & 10 training iterations\\
PCA dimension / solver & 64 / randomized\\
K-means \texttt{n\_init} & 512\\
\bottomrule
\end{tabular}
\end{table}

This subsection supplements the common settings above with method-specific configurations for the baselines and ASE-GRPO. The two tables summarize the policy optimization settings and extra processing used by each method, together with the activation readout, routing, steering, and vector-bank update settings used by ASE-GRPO.
For rollout correction, we use verl's default token-level truncated importance-sampling configuration, with $c_{\mathrm{IS}}=2.0$ and no batch normalization of the weights.
For layer choice, we read out activations and apply steering at the output of Transformer block 0 in all ASE experiments. We use a single intervention layer to avoid accumulating direct vector additions across multiple depths, while an early intervention allows the injected direction to propagate through the remaining blocks. This choice is heuristic: intermediate or later layers may provide more informative representations for constructing exploration directions. We did not run a layer sweep within the available compute budget, so we do not claim that block 0 is optimal; comparing readout and intervention layers, including multi-layer configurations, remains future work.

\section{Additional Ablation}
\label{app:additional-ab}

\subsection{Hyperparameter Sensitivity}
\label{app:hyperparameter-sensitivity}

We report ASE sensitivity sweeps for the number of clusters $K$, the EMA coefficient $\beta$, the steering strength $\alpha$, and the response-relative activation readout position $p$. Each run uses its best validation checkpoint. Entries are Mean@8 / Pass@8 percentages; the mathematical average covers the six mathematical benchmarks and the OOD average covers MMLU-Pro, GPQA-D, and BBH. The sweeps vary one parameter at a time, with the remaining parameters fixed as indicated in each table.

\begin{table}[h]
\centering\small
\caption{Sensitivity to the number of clusters $K$. For this ASE sweep, we fix $\alpha=0.4$ and $\beta=0.5$.}
\label{tab:ablation-k}
\begin{tabular}{ccc}
\toprule
$K$ & Math Avg. & OOD Avg. \\
\midrule
2 & 29.60/45.74 & 40.08/78.44 \\
3 & 29.64/47.02 & 39.62/76.52 \\
4 & 27.01/44.63 & 39.40/78.24 \\
5 & 28.04/46.38 & 38.82/78.26 \\
6 & 28.04/44.54 & 39.54/77.72 \\
\bottomrule
\end{tabular}
\end{table}

\begin{table}[h]
\centering\small
\caption{Sensitivity to the EMA coefficient $\beta$. For this ASE sweep, we fix $K=3$ and $\alpha=0.2$.}
\label{tab:ablation-ema}
\begin{tabular}{ccc}
\toprule
$\beta$ & Math Avg. & OOD Avg. \\
\midrule
0.0 & 26.48/42.63 & 35.84/72.25 \\
0.1 & 28.29/45.41 & 38.02/72.10 \\
0.3 & 26.27/46.10 & 35.33/76.78 \\
0.5 & 30.21/46.44 & 40.79/79.71 \\
0.7 & 29.41/48.83 & 37.14/79.14 \\
0.9 & 29.35/47.81 & 39.89/78.35 \\
\bottomrule
\end{tabular}
\end{table}

\begin{table}[h]
\centering\small
\caption{Sensitivity to the steering strength $\alpha$. For this ASE sweep, we fix $K=3$ and $\beta=0.5$.}
\label{tab:ablation-strength}
\begin{tabular}{ccc}
\toprule
$\alpha$ & Math Avg. & OOD Avg. \\
\midrule
0.2 & 30.21/46.44 & 40.79/79.71 \\
0.3 & 28.38/48.05 & 38.75/77.62 \\
0.4 & 29.64/47.02 & 39.62/76.52 \\
0.5 & 30.31/48.67 & 41.35/79.97 \\
0.6 & 29.21/44.63 & 39.43/76.84 \\
\bottomrule
\end{tabular}
\end{table}

The three sweeps suggest that a small route bank is sufficient under the
tested rollout budget. For the cluster-count sweep, $K=3$ gives the highest
mathematical Mean@8 and Pass@8, at 29.64 and 47.02, respectively, whereas
$K=2$ gives the highest OOD Mean@8 and Pass@8, at 40.08 and 78.44. Increasing
$K$ beyond three does not produce a monotonic improvement: both mathematical
metrics are lower for $K\in\{4,5,6\}$ than for $K=3$. This indicates that
adding routes alone does not guarantee better coverage, and that the
available rollout evidence must also support reliable estimation of each
direction.

The EMA sweep supports using temporal smoothing when updating the vector bank.
Relative to the unsmoothed setting $\beta=0$, the intermediate setting
$\beta=0.5$ improves mathematical Mean@8 from 26.48 to 30.21 and OOD Mean@8
from 35.84 to 40.79. It also gives the highest OOD Pass@8, 79.71, while the
highest mathematical Pass@8 is obtained at $\beta=0.7$, with 48.83. Thus,
moderate-to-strong smoothing performs well across the reported metrics, with
$\beta=0.5$ providing a useful overall operating point rather than being the
strict optimum for every metric.

The strength sweep shows non-monotonic variation across the tested values of
$\alpha$. Mathematical Mean@8 ranges from 28.38 to 30.31 and Pass@8 from
44.63 to 48.67; OOD Mean@8 ranges from 38.75 to 41.35 and Pass@8 from
76.52 to 79.97. All four metrics reach their highest values at $\alpha=0.5$:
30.31/48.67 for Math Avg. and 41.35/79.97 for OOD Avg. Increasing the
strength to $\alpha=0.6$ lowers all four metrics, so stronger steering does
not consistently improve the final unsteered policy over this range.

These tables are separate one-parameter sensitivity analyses rather than a
joint factorial study. In particular, the cluster-count sweep fixes
$\alpha=0.4$, the EMA sweep fixes $\alpha=0.2$, and the strength sweep varies
$\alpha$ with $K=3$ and $\beta=0.5$. The different fixed strengths reflect
separate experimental runs that inherited the working configuration used at
the time of each sweep, rather than a rerun of every sweep after selecting a
single global anchor. This design keeps each table's within-sweep comparison
well-defined, but it prevents comparisons of absolute values across tables
from being interpreted as a unified hyperparameter ranking. In particular,
the best row in each table should not be combined to claim a jointly optimal
$(K,\beta,\alpha)$ configuration. A fully matched sensitivity study would
hold one common anchor fixed for all three scans, or would evaluate a
prespecified joint grid; the present results are intended to characterize
local sensitivity around the configurations used in the respective runs.

We also vary the response-relative readout position on Qwen3-1.7B-Base in a
separate sweep, selecting activations at a fraction $f$ of each response's
length. Its absolute scores are not directly comparable with those of the
preceding ablations since the sweeps were not conducted under a common matched
configuration. Within this sweep, $f=0.25$ performs best on the mathematical
benchmarks, the primary target of ASE. One possible explanation is that earlier
response states reflect the initial solution strategy before later
calculations and answer formatting add trajectory-specific variation, making
success--failure contrasts more reusable across problems. This explanation
remains a hypothesis based on single-run comparisons.

\begin{table}[h]
\centering\small
\caption{Sensitivity to the response-relative activation readout position.
We fix $K=3$, $\alpha=0.2$, and $\beta=0.7$; scores are Mean@8 / Pass@8
(\%). Math Avg. and OOD Avg. use the same benchmark groups as the main tables.}
\label{tab:ablation-readout}
\begin{tabular}{ccc}
\toprule
Readout fraction $f$ & Math Avg. & OOD Avg. \\
\midrule
0.25 & 28.93/46.79 & 38.24/77.59 \\
0.50 & 27.89/43.94 & 39.68/76.98 \\
0.75 & 27.59/44.95 & 40.00/78.36 \\
\bottomrule
\end{tabular}
\end{table}

\subsection{Ablating Verifier-Guided Direction}
\label{app:matched-direction}
The matched controls in Table~\ref{tab:matched-direction} change direction construction while retaining the clustered $K=3$ bank, routing, and EMA updates. For random directions, each independent random vector is rescaled to the corresponding cluster candidate's original $L_2$ norm before the EMA update. For shuffled labels, the positive and negative labels within each activation pair are randomly exchanged before PCA, K-means, and cluster-direction estimation. Consequently, the controls' gains over GRPO do not isolate the effect of arbitrary activation perturbations: the shared bank structure and temporal averaging may also contribute.

To examine this possibility, we additionally use unstructured random steering: eight sampled unit random vectors, one per rollout, with no clustering or EMA. The vectors do not depend on activations or verifier labels. As shown in Table~\ref{tab:unstructured-random-bank}, this control reaches 26.65/41.59 on Math Avg., compared with 26.00/43.91 for GRPO and 30.31/48.67 for ASE-GRPO. Its Mean@8 is only 0.65 points above GRPO, while its Pass@8 is 2.32 points lower, so arbitrary random steering does not yield a consistent improvement in this comparison. This pattern is consistent with the structured bank and EMA contributing to the gains of the matched controls.

\begin{table}[t]
\centering\small
\caption{Six-benchmark Math Avg. (Mean@8 / Pass@8, \%). GRPO and ASE-GRPO use the first evaluation at their selected checkpoints.}
\label{tab:unstructured-random-bank}
\begin{tabular}{lc}
\toprule
Method & Math Avg. \\
\midrule
GRPO & 26.00/43.91 \\
Unstructured random steering & 26.65/41.59 \\
ASE-GRPO & 30.31/48.67 \\
\bottomrule
\end{tabular}
\end{table}

\subsection{Inference-time steering sweep}
\label{app:inference-sweep}
The sweep compares three checkpoints: Qwen3-1.7B-Base and its GRPO- and ASE-GRPO-trained variants. For each model, vectors are constructed from response-relative activations captured on the 17,390 filtered DAPO-Math-17k prompts. We use block 0, within-problem verifier-reward pairing, CAA-style positive-minus-negative differences, no cluster-feature scaling, PCA dimension 64, K-means \texttt{n\_init}=512, balanced-uniform route allocation, and a clean route. We evaluate
\begin{equation}
K\in\{1,2,3,4,5,6,7\},\qquad
f\in\{0.25,0.50,0.75\},\qquad
\alpha\in\{0.25,0.50,0.75,1.00\}.
\end{equation}
This gives 84 steering conditions per model. Each steering condition is independently repeated three times, with eight responses per problem. The Base and GRPO clean references use three standalone evaluations; the ASE-GRPO clean reference combines its selected-checkpoint training validation with two standalone evaluations. Evaluation uses MATH500, Minerva, AIME24, AIME25, and AMC24, temperature 0.6, top-$p$ 0.95, top-$k=0$, maximum response length 4096, and thinking mode.

For a sweep cell, Core Math Mean@8 and Pass@8 are the unweighted averages across the five benchmarks. We report the mean and sample standard deviation over the three evaluation repeats. The corresponding clean-relative changes use the same model's clean mean:
\begin{equation}
\Delta_{\mathrm{mean}}=
\overline{\mathrm{Mean@8}}_{\mathrm{steer}}-
\overline{\mathrm{Mean@8}}_{\mathrm{clean}},\qquad
\Delta_{\mathrm{pass}}=
\overline{\mathrm{Pass@8}}_{\mathrm{steer}}-
\overline{\mathrm{Pass@8}}_{\mathrm{clean}}.
\end{equation}
Table~\ref{tab:inference-steering} uses the highest Core Math Mean@8 configuration from each model's sweep: $(K,f,\alpha)=(7,0.50,0.50)$ for Base, $(7,0.25,0.25)$ for GRPO, and $(4,0.75,0.50)$ for ASE-GRPO. These sweep-selected results are interpreted as exploratory upper envelopes.

\begin{table}[h]
\centering\small
\caption{Inference-time sweep grid and reporting units.}
\label{tab:sweep-grid}
\begin{tabular}{ll}
\toprule
Quantity & Values or definition \\
\midrule
Models & Base; standard GRPO; ASE-GRPO \\
Number of routes $K$ & $1,2,3,4,5,6,7$ \\
Capture fraction $f$ & $0.25,0.50,0.75$ \\
Strength $\alpha$ & $0.25,0.50,0.75,1.00$ \\
Conditions per model & $7\times3\times4=84$ \\
Repeats & Three independent evaluations per cell \\
Responses & Eight per problem \\
Core Math Avg. & Unweighted average over five benchmarks \\
Selected Base $(K,f,\alpha)$ & $(7,0.50,0.50)$ \\
Selected GRPO $(K,f,\alpha)$ & $(7,0.25,0.25)$ \\
Selected ASE-GRPO $(K,f,\alpha)$ & $(4,0.75,0.50)$ \\
\bottomrule
\end{tabular}
\end{table}

\end{document}